# Cascaded Semantic and Positional Self-Attention Network for Document Classification

**Juyong Jiang**[1], **Jie Zhang**[2], **Kai Zhang**[3*]
[1]College of Internet of Things Engineering, Hohai University, Nanjing, China
[2]Institute of Science and Technology for Brain-Inspired Intelligence, Fudan University, Shanghai, China
[3]Department of Computer & Information Sciences, Temple University, PA, USA
[1]jiangjuyong@hhu.edu.cn
[2]jzhang080@gmail.com, [3]zhang.kai@temple.edu

## Abstract

Transformers have shown great success in learning representations for language modelling. However, an open challenge still remains on how to systematically aggregate *semantic* information (word embedding) with *positional (or temporal)* information (word orders). In this work, we propose a new architecture to aggregate the two sources of information using cascaded semantic and positional self-attention network (CSPAN) in the context of document classification. The CSPAN uses a semantic self-attention layer cascaded with Bi-LSTM to process the semantic and positional information in a sequential manner, and then adaptively combine them together through a residual connection. Compared with commonly used positional encoding schemes, CSPAN can exploit the interaction between semantics and word positions in a more interpretable and adaptive manner, and the classification performance can be notably improved while simultaneously preserving a compact model size and high convergence rate. We evaluate the CSPAN model on several benchmark data sets for document classification with careful ablation studies, and demonstrate the encouraging results compared with state of the art.

## 1 Introduction

Document classification is one of the fundamental problems in natural language processing, which is aimed at assigning one or multiple labels to a (typically) short text paragraph. Wide applications can be found in sentiment analysis (Moraes et al., 2013; Tang et al., 2015)，subject categorization (Wang et al., 2012), spam email detection (Sahami et al., 1998) and doc[1]ument ranking (Wang et al., 2014). In recent years, deep neural networks have shown great potential in document classification and updated state-of-the-art performance. Popular approaches include Recurrent neural networks (RNN) (Yogatama et al., 2017), convolutional neural networks (CNN) (Zhang et al., 2015) and Attention-based methods (Transformers) (Gong et al., 2019; Adhikari et al., 2019), or a mixture of them.

Different lines of methods have their respective pros and cons. For example, RNNs are highly effective models for exploiting word orders in learning useful representations, thanks to the iterative update of the hidden states that depend on both the semantics of the current word and that of historical words (or a concise summary of them), and the long-range dependency made possible through LSTMs (Yang et al., 2016; Stephen et al., 2018; Adhikari et al., 2019). Of course, the sequential processing nature makes it less efficient computationally. CNNs have gained huge success in image procesing and classification and were recently introduced to NLP domains like document classification (Zhang et al., 2015; Lei et al.,2015; Conneau et al., 2016; Kim and Yang, 2018; Kim, 2014). The local convolutional operator is sensitive to word orders but only partially and limited by the size of the kernel, and so long-term relations may need many layers and therefore be challenging. Transformers, different from both, fully exploit the modelling power of self-attention mechanism (Shen et al., 2018; Gao et al., 2018; Zheng et al., 2018) and have significantly improved state of the art in many NLP tasks such as machine translation (Vaswani et al., 2017),

---
[*] Corresponding author.

language understanding (Devlin et al., 2018) and language modeling (Dai et al., 2019), etc.

Despite the great successes, how to systematically aggregate the *semantic* information (word embedding) with the *positional* information (word orders) is still an open challenge in transformers. A common practice is the positional encoding (Vaswani et al., 2017), which encodes the position of the $t$th word as a $d$-dimensional sinusoidal vector, as

$$p_{t,2i} = sin(t/10000^{2i/d}), \quad (1)$$

$$p_{t,2i+1} = cos(t/10000^{2i/d}). \quad (2)$$

The positional vector of each word is then added to the $d$-dimensional word embedding vector, so that subsequent predictors can numerically utilize the temporal information. However, empirically, adding positional vectors to the word vectors brings little performance gains in document classification, compared with when no positional encoding is adopted at all (See Section 3.4 Table 5 for detailed empirical results).

There are two reasons which we believe are related to the low performance gains from using positional encodings. First, such a strategy leads to an interaction (inner product) between the semantic and temporal component that is hard to interpret. To see this, let $x_i$ and $p_i$ be the word vector and position vector for the $i$th word. Then the attention score between $i$th and $j$th word will be computed as (before normalization)

$$\begin{aligned} e_{ij} &= \langle x_i + p_i, x_j + p_j \rangle \\ &= \langle x_i, x_j \rangle + \langle p_i, p_j \rangle + \langle x_i, p_j \rangle \\ &\quad + \langle p_i, x_j \rangle \end{aligned} \quad (3)$$

where $\langle \cdot, \cdot \rangle$ denotes the inner product between two vectors, and without loss of generality we have assumed identity transforms in generating the key and query views of each word.

Obviously, as the inner product between a word vector and a positional vector, $\langle x_i, p_j \rangle$ and $\langle p_i, x_j \rangle$ do not bear meaningful interpretation. Therefore these two terms could very likely hamper the semantic attention term $\langle x_i, x_j \rangle$ and the positional attention term $\langle p_i, p_j \rangle$ by behaving like noise, such as deflating an important attention or exaggerating a marginal one. This can negatively affect the learned representations through the self-attention mechanism. Indeed, similar observations were made in (Yan et al., 2019), where the authors show that the self-attention mechanism, when mixed with the positional vectors, can no longer effectively quantify the relative positional distance between the words (namely the positional attention term $\langle p_i, p_j \rangle$ is perturbed in an undesired manner).

Second, the relative weights of the word vector and the position vector (in their summation) is hard-coded, leading to a fixed combination, while in practice the relative importance of the semantic and positional components in affecting the similarity among the words can definitely be more complex.

In order to solve these challenges with positional encoding, we explore a new architecture in combining the semantic and temporal information in document classification, called "cascaded semantic and positional self-attention network" (CSPAN). There are three main characteristics of the proposed architecture. First, instead of combining the word vectors with positional vectors from scratch, we choose to first explore the two sources of information with their respective processing layers, namely, a self-attention layer that works only on the semantic space, and a Bi-LSTM layer which further incorporates the temporal order information in the updated word representations. Second, these two layers are cascaded so that sematic information and the temporal information can be finally combined through the use of a residual connection; this not only avoids non-interpretable operations defined between word vectors and positional vectors, but also serves as an adaptive transformation in combining the two information sources. Third, a multi-query attention scheme is adopted to extract multi-faceted, fixed dimensional document features, which makes the resultant model highly compact and memory efficient.

The CSPAN model is shown to effectively improve performance of document classification in comparison to several state-of-the-art methods including transformer-styled architecture. In the meantime, it demonstrates very compact model size and fast convergence rate during the training process, which is particularly desirable for large problems. We also conducted careful ablation studies to further quantify the performance gains of each component of the CSPAN model.

Our study demonstrates the importance of the way semantic and temporal information are aggregated in capturing the structures and meaning of documents, which we will continue exploring in the more challenging language modelling tasks such as sequence tagging (Huang et al., 2015), natural language inference (Chen et al., 2016) and modeling sentence pairs (Tan et al., 2018) in our future research.

## 2 Method

The overall architecture of the proposed CSPAN model is shown in Figure 1. It is a highly compact model with three basic building blocks.

First, we use a self-attention block to update the word representations in each document. Here, the embedding of each word will be collectively affected by all other words with related semantics in the same document. Note that we will not look into any positional information in this stage. Instead, the temporal information will be taken into account in the next block, after the word representations have been fully updated through semantic self-attention alone. As we shall see, such a sequential processing pipeline allows more flexible combination of the semantic and positional information.

Second, the updated word embeddings are fed into a Bi-LSTM layer, so that the relative position of the words are naturally exploited to further refine the word representations specific to the organization of each document. In the meantime, a residual connection is adopted to combine the semantic representation derived from the self-attention block, together with the output derived from the Bi-LSTM block; we call this ``Semantic and Positional Residual Connection'', because it combines the semantic information (out of self-attention block) with the positional information (out of the Bi-LSTM block) using residual connections. As we shall see, such a combination is more flexible than directly combining word vector with positional vector as in existing positional encoding schemes.

Third, we adopt a multiple-query attention in the final block to extract fixed-dimensional document features for final classification. Compared with multi-head attention, the multi-query attention can significantly reduce the number of parameters in the network, while giving promising classification results. We describe the details of different structures and components of our model in the following sections.

### 2.1 Semantic Self-Attention

Self-attention as proposed by (Vaswani et al., 2017)

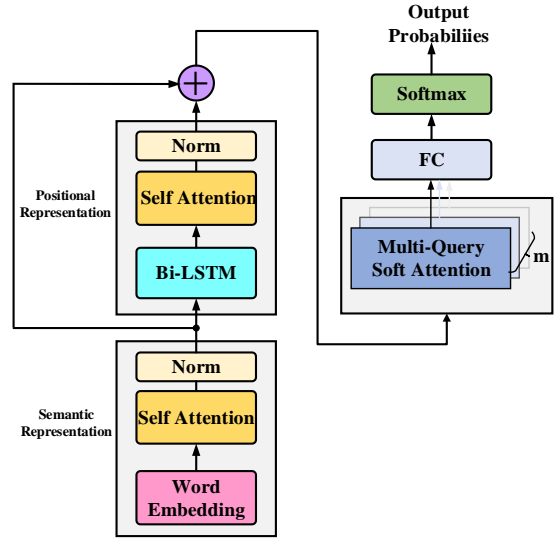

Figure 1: The architecture of the proposed CSPAN model.

calculates attention weight between each pair of objects to capture global correlations and improve representation learning. We apply this framework in computing the word representations since it can capture long-range dependencies. However, we do make a number of important rectifications which prove to be quite useful in improving the performance of document classification.

First, rather than using three independent transformation matrices corresponding to the key, value, and query views for each word, we discard these transformations, and use the original word vectors in all the three views. The reason is that we want to activate a full, pairwise interaction between the words in the original word embedding space and then apply transformations in subsequent (Bi-LSTM) layer, in order to maximally preserve the power of self-attention based representation learning. In comparison, if one chooses to apply transformation (e.g. dimensionality reduction in most cases), then chances are that the semantic information encoded in the word vectors might suffer certain losses before entering the next layer. Empirically, we have observed that implementing self-attention in the full-dimensional word vectors leads to better performance than that on the lower dimensional, transformed word-vectors.

Second, rather than considering the use of the positional information in self-attention, we choose to implement self-attention only based on the semantic information, and consider the positional information in subsequent information processing blocks. This in contrast to current practices in which the semantic information and positional information of each word is used together in calculating the self-attention coefficients. The reason is that directly adding the word vector and positional vector can lead to noisy fluctuations in attention scores, as has been discussed in the introduction. Therefore, the semantic information will first be processed alone, and then subject to the positional information through subsequent LSTM layer, which is a more natural way of injecting positional information.

Given these two design principles, our self-attention block can be described as follows. Let the input text sequence be $D = (w_1, w_2, ..., w_L)$ of $L$ elements where $w_i \in \mathbb{R}^d$ is the i-th word embedding. Self-attention compares each element $w_i$ to every other element $w_j$ in the sequence followed by layer normalization. As a result, a new sequence $S = (s_1, s_2, ..., s_L)$ of the same length is constructed, in which each element $s_i \in \mathbb{R}^d$ is a weighted average of all elements $w_i$ in the input sequence, as

$$S = Attention(D, D, D)$$
$$= softmax\left(\frac{DD^T}{\sqrt{d}}\right) D \quad (4)$$

Here, the original word embedding matrix $D \in \mathbb{R}^{L \times d}$ appears three times because we do not differentiate among the key, value and query views. The term $DD^T$ is used to generate a weight matrix based on the inner-product similarity of the elements in the sequence. After normalization and re-scaling, the weight matrix is multiplied with $D$ to generate the new sequence representation $S$. The self-attention can enhance the semantic representation of word embeddings and capture both the local and long-range dependencies.

## 2.2 Semantic and Positional Residual Connection

In the second block, we apply a Bi-LSTM layer to inject temporal information in the word representations computed via the self-attention block. The Bi-LSTM is a powerful model in handling sequential data, and is known to capture long-term dependencies due to the use of the gating mechanism (Graves and Schmidhuber, 2005). Therefore this layer is supposed to further improve the word representations obtained from the self-attention layer, which proceeds as

$$\vec{h}_t = \overrightarrow{LSTM}(s_t) \quad (5)$$
$$\overleftarrow{h}_t = \overleftarrow{LSTM}(s_t) \quad (6)$$

$$h_t = [\vec{h}_t, \overleftarrow{h}_t] \quad (7)$$
$$P = Attention(H, H, H) \quad (8)$$

Here, the word vectors obtained through the self-attention layer, $s_i's \in \mathbb{R}^d$ are fed into a single-layer Bi-LSTM, and then the hidden state of the LSTM in the forward and backward directions are concatenated as $h_t = [\vec{h}_t, \overleftarrow{h}_t]$. Finally, another self-attention layer is used to enhance the representations $H = [h_1, h_2, ..., h_L]$, followed by a layer-wise normalization to obtain the position-aware representations $P = (p_1, p_2, ..., p_L)$.

Although LSTMs are known to handle long-range dependencies, it can still be challenging in long documents. Therefore, following the custom in transformers (Vaswani et al., 2017), we use a residual connection that combines the output of the self-attention layer with that of the Bi-LSTM layer, computed as shown below.

$$F_t^{sp} = s_t + p_t \quad (9)$$

Here, $s_t \in \mathbb{R}^d$ represents the output of first building block (Semantic self-attention), $p_t \in \mathbb{R}^d$ stands for the output of second building blocks (Bi-LSTM). To guarantee that the two vectors can be added together, the hidden-state dimension of the Bi-LSTM is chosen as half of the input dimension, i.e., $d/2$, so that the concatenated hidden state from the forward and backward direction (7) has the same dimension as the input word vectors. By combining the semantic and positional information, we obtain a final, high-level representation of each document.

The residual connection (He et al., 2016) has shown to be highly useful in facilitating an effective backpropagation so that the learning process approaches a better model. In our context, the residual connection has an interesting interpretation of combining sematic and positional information in an adaptive manner. Note that the output of the self-attention layer is all about the semantic component of the words; on the other hand, the output of the Bi-LSTM layer can be deemed as word representations that incorporated the positional information, thanks to the sequential processing nature of the Bi-LSTM. Besides, since the output of the Bi-LSTM layer, its hidden state, is a transformation of the input word vectors, we can then consider the output of the residual connection as an adaptive combination of the semantic components and positional components. This not only avoids the non-interpretability of directly combining word vector with position vectors, but also successfully adjusts their relative importance through the learning of the transformation matrices in the Bi-LSTM model. We speculate that this is an important reason why the proposed architecture can effectively improve the classification performance.

### 2.3 Multi-Query Soft Attention

In the final block, we learn a number of query vectors in the space of $F_t^{sp}$ (9) so that each query can capture a certain aspect of the meaning of the document, in the form of a fixed-dimensional feature (context) vector. This is in contrast to the single-query attention where only a single query vector is learned to summarize the content of a document (Yang et al., 2016). It is worthwhile to note that the multi-query attention in extracting document features can be computationally more effective than multi-head attention. In the latter case, one attention head is associated with a independent set of transformation matrices, therefore the model size can be quite large. In comparison, in our approach only multiple query vectors need to be learned in the same latent space of word representations, which has a much smaller memory footprint.

More Specifically, the multi-query attention is defined as follows.

$$u_t = tanh(F_t^{sp} W^h + b^h) \quad (10)$$

$$\alpha_{it} = \frac{exp(u_t^T Q_i)}{\sum_t exp(u_t^T Q_i)} \quad (11)$$

$$F_i^{spmq} = \sum_t \alpha_{it} F_t^{sp} \quad (12)$$

$$\tilde{F}^{spmq} = Concat(F_1^{spmq}, \dots, F_m^{spmq}) W^f \quad (13)$$

That is, we first feed the $F_t^{sp} \in \mathbb{R}^d$ through a one-layer MLP to get $u_t \in \mathbb{R}^d$ as a hidden representation of $F_t^{sp} \in \mathbb{R}^d$, then we measure the importance of the word as the similarity of $u_t$ with a query vector $Q_i \in \mathbb{R}^d$ and get a normalized importance weight $\alpha_i \in \mathbb{R}^L$ through a softmax function. The multi-query matrix is randomly initialized and jointly learned during the training process. After that, we compute the $F_i^{spmq} \in \mathbb{R}^d$ as a weighted sum of the $F_t^{sp} \in \mathbb{R}^d$ based on the weighting. Finally, we concatenate all $F_i^{spmq}$ vectors and then use a fusion matrix $W^f \in \mathbb{R}^{md \times d}$ to get a high-level representation of each document.

Here we discuss in more detail the memory footprint of the proposed multi-query attention, in comparison and commonly used multi-head attention. Let the dimension of the residual connection be $d$; the number of query vectors be $m$. Then the model space complexity is $O(md + d^2)$. In comparison, if one adopts the multi-head attention with $m$ attention heads, then the model space complexity will be $O(md^2)$ since each attention head will have its own transformation parameters. As can be seen, the memory saving is almost proportional to the dimensionality; the higher the word vector dimensions, the more significant the memory saving. This will be a desired property for real-world applications. It is also worthwhile to note that the CSPAN model only has 3 blocks, while the standard transformer has a cascade of 6 layers of self-attention each of which may require an independent set of transformation matrices.

### 2.4 Classification Layer

In the final layer we apply a softmax classifier on the document representation $\tilde{F}^{spmq}$ to get a predicted label $\hat{y}$, where $\hat{y} \in Y$ and $Y$ is the class label set, i.e.,

$$\hat{y} = argmax \, p(Y|\tilde{F}^{spmq}) \quad (14)$$

where

$$p(Y|\tilde{F}^{spmq}) = softmax(W^o \tilde{F}^{spmq} + b^o) \quad (15)$$

Here, $W^o$ and $b^o$ are the transformation matrix and the bias term, respectively. Therefore, we can use the negative log-likelihood to define the loss function as follows:

$$L = -\log p(\hat{y}|\tilde{F}^{spmq}) \quad (16)$$

## 3 Experiments

In this section, we will report a number of experimental results on 4 benchmark datasets for document classification, together with careful ablation studies to illustrate the effectiveness of the building blocks of the proposed method.

### 3.1 Datasets and Methods

We evaluate the effectiveness of the proposed CSPAN model on four document classification datasets as in (Zhang et al., 2015). The detailed statistics of the data sets are shown in Table 1.

**AG's News.** Topic classification over four categories of internet news articles composed of titles plus description classified into: World, Sports, Business and Sci/Tech. The number of training samples for each class is 30,000 and testing 1900.

| Dataset | Classes | Train | Test | Average #s | Max #s | Average #w | Max #w |
|---|---|---|---|---|---|---|---|
| AG's News | 4 | 120,000 | 7,600 | 1.3 | 15 | 46.6 | 277 |
| Yelp Review Polarity | 2 | 560,000 | 38,000 | 8.4 | 119 | 161.4 | 1345 |
| Yelp Review Full | 5 | 650,000 | 50,000 | 8.4 | 151 | 163.3 | 1418 |
| Yahoo! Answers | 10 | 1,400,000 | 60,000 | 5.7 | 515 | 115.9 | 2746 |

Table 1: Detailed statistics of the datasets: #s denotes the number of sentences (average and maximum per document), #w denotes the number of words (average and maximum per document).

**Yelp Review Polarity.** The same dataset of text reviews from Yelp Dataset Challenge in 2015, except that a coarser sentiment definition is considered: 1 and 2 are negative, and 4 and 5 as positive. The polarity dataset has 280,000 training samples and 19,000 test samples in each polarity.

**Yelp Review Full.** The dataset is obtained from the Yelp Dataset Challenge in 2015 on sentiment classification of polarity star labels ranging from 1 to 5. The full dataset has 130,000 training samples and 10,000 testing samples in each star.

**Yahoo! Answer.** Topic classification over ten largest main categories from Yahoo Answers Comprehensive Questions and Answers version 1.0: Society & Culture, Science & Mathematics, Health, Education & Reference, Computers & Internet, Sports, Business & Finance, Entertainment & Music, Family & Relationships and Politics & Government. The document we use includes question titles, question contexts and best answers. Each class contains 140,000 training samples and 5,000 testing samples.

**Methods.** We have included altogether eleven competing methods from (Zhang et al., 2015) and (Gong et al., 2019). For our approach, we have two versions: the CSPAN (base) using single-layer Bi-LSTM and 16 query vectors, and CSPAN (big) using three hidden layers in Bi-LSTM and 128 query vectors. We trained the base models for 30 epochs and the big models for 60 epochs.

### 3.2 Model configuration and training

In the experiments, we use 300-dimensional GloVe 6B pre-trained word embedding (Pennington et al., 2014) to initialize the word embedding at https://nlp.stanford.edu/projects/glove. We choose 150 hidden units for the Bi-LSTM models. The Adam Optimizer (Kingma et al., 2014) with learning rate of 1e-3 and weight decay of 1e-4 is used to train the model parameters. The size of mini-batch is set to 64 and the number of multi-query to 16. We train all neural networks for 30 epochs and the learning rate divides by 10 at 20 and 25 epochs. All of our experiments are performed on NVIDIA TITAN RTX GPUs, with PyTorch 1.1.0 as the backend framework.

### 3.3 Results and analysis

The experimental results on all data sets are shown in Table 2. The results of the competing methods are directly cited from the respective papers as listed in Table 2.

From Table 2 we can see that CSPAN model achieves the best performance on all the 4 datasets of AG's News, Yelp P, Yelp F. and Yahoo datasets (rows 12/ 13), which demonstrates its effectiveness in document classification. Particularly, CSPAN consistently outperforms the baseline deep learning networks using RNN/CNN, such as LSTM, CNN-char and CNN-word by a substantial margin on all datasets (rows 1, 2 and 3).

Compared to the CSPAN (base), the CSPAN (big) gives a comparable or slightly better performance on all the datasets. This observation shows that the CSPAN actually prefers simpler models against highly complex ones, which is an advantage for large problems.

### 3.4 Ablation Study

**Component-wise gains.** To investigate the impact of each of the key components of CSPAN model for document classification, we conducted an ablation study on the AG's News dataset. Firstly, we validate the impact of each component, including semantic self-attention, semantic and positional residual connection, and multi-query soft attention. The results are shown in Table 3.

The standard Bi-LSTM baseline provides a test accuracy of 89.36. As we expected, integrating semantic self-attention significantly improved the classification performance with test accuracy of 92.61. It shows that using self-attention can enhance the semantic. Furthermore, integrating residual connection improves the classification performance from 92.61 to 93.03. Finally, when multi-query attention is adopted, the classification performance is significantly improved with an overall gain of 4.32% over the baseline.

|  | Methods | AGNews | Yelp P. | Yelp F. | Yahoo |
|---|---|---|---|---|---|
| Zhang et al., 2015 | LSTM | 86.06 | 94.74 | 58.17 | 70.84 |
|  | CNN-char | 89.13 | 94.46 | 62.02 | 69.98 |
|  | CNN-word | 91.45 | 95.11 | 60.48 | 70.94 |
| Gong et al., 2019 | Deep CNN | 91.27 | 95.72 | 64.26 | 73.43 |
|  | FastText | 92.50 | 95.70 | 63.90 | 72.30 |
|  | HAN | 92.36 | 95.59 | 63.32 | 75.80 |
|  | SASEM | 91.50 | 94.90 | 63.40 | - |
|  | DiSAN | 92.51 | 94.39 | 62.08 | 76.15 |
|  | LEAM | 92.45 | 95.31 | 64.09 | 77.42 |
|  | SWEM | 92.24 | 93.76 | 61.11 | 73.53 |
|  | HLAN | 92.89 | 95.83 | 63.78 | 77.55 |
| This paper | CSPAN (base) | **93.68** | 96.11 | 65.93 | 77.61 |
|  | CSPAN (big) | 93.62 | **96.18** | **65.95** | **77.75** |

Table 2: Test accuracy of competing methods on benchmark document classification tasks, in percentage.

| Component | Accuracy |
|---|---|
| Standard Bi-LSTM(baseline) | 89.36 |
| + self-att | 92.61 |
| + residual | 93.03 |
| + multi-query | **93.68** |

Table 3: Impact of each building block in the proposed CSPAN model on AG's News dataset.

| Layers (BiLSTM) | Query | Memory(MB) | Accuracy |
|---|---|---|---|
| 1 | 1 | 1557 | 92.84 |
| 1 | 8 | 1641 | 92.95 |
| 1 | 16 | 1739 | **93.68** |
| 2 | 8 | 1665 | 93.05 |
| 2 | 16 | 1765 | 92.88 |
| 2 | 32 | 1961 | 93.04 |
| 3 | 32 | 1997 | 92.92 |
| 3 | 64 | 2401 | 92.71 |
| 3 | 128 | **3201** | 93.14 |

Table 4: Impact of model size.

| # | Methods | Accuracy |
|---|---|---|
| (a) | Embedding | 92.38 |
| (b) | Embedding + Position | 92.71 |
| (c) | Embedding + Relative-Position | 92.39 |
| (d) | Embedding + Bi-LSTM | 93.03 |
| (e) | Embedding // Bi-LSTM | **93.68** |

Table 5: Different ways in combining the semantic and the positional information and their accuracy on AG's News dataset.

**Model Size.** As mentioned in (Adhikari et al., 2019), increasingly complex network components and modeling techniques are accompanied by smaller and smaller improvements in effectiveness on standard benchmark datasets. We have observed similar trend in CSPAN, as shown in Table 4.

From Table 4, we can see that when the number of hidden layer in Bi-LSTM is set to 3, the performance can be worse than 1-layer or 2-layer Bi-LSTMS (the latter with even less query vectors). In other words, a compact Bi-LSTM is preferred. On the other hand, the optimal number of query vectors seems to be around 16 for 1-layer Bi-LSTM; more query vectors than this brings limited or even negative performance gains.

**Fusion Methods.** We also conducted extensive comparative studies on the performance of different ways in combining the semantic and the positional information, as shown in Figure 2.

From Table 5, we can see that directly combining the positional vector with the word vector (fusion method (b), a "light-weight" transformer) brings an improvement of 0.33% compared with the baseline (method (a), without any positional information). In addition, using relative positional encoding schemes (Shaw et al., 2018) (fusion method (c)) leads to almost the same result as the baseline method. If we use Bi-LSTM directly on the input word vectors, i.e., a parallel combination scheme of the semantic and positional information (fusion method (d)), the performance gain approaches 0.65%. Finally, the proposed fusion scheme in CSPAN (fusion method (e)), i.e., sequential processing of semantic and positional information equipped with a residual connection, the performance gain is around 1.30%. This comparative study clearly demonstrates the advantage of the proposed CSPAN model in combining semantic and positional information.

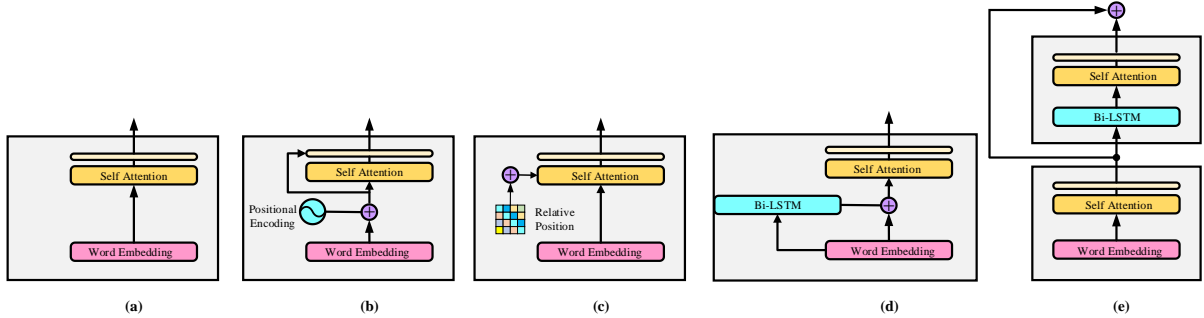

Figure 2: Different schemes of combining the semantic and position information for a comparative study, where (b) corresponds to a "light-weight"-transformer, and (e) is the proposed architecture.

**Computational Considerations.** It is usually believed that transformers are computationally efficient by virtue of the parallel processing pipeline associated with the self-attention mechanism. However, empirically, we find that the large model size and extensive, pairwise self-attention cost can significantly slow down the computation. For example, standard transformers have 6 layers of self-attention in the encoding stage alone, leading to a huge set of transformation matrix parameters $W^Q, W^K, W^V$ and the cost of back-propagation can be huge. On the other hand, $O(n^2)$ time and space are needed in each layer in computing the self-attention among a document of $n$ words. Therefore, standard transformer is time consuming in our experimental evaluations and typically won't converge until after tens or even 100 epochs even on the smallest data set (AG's News). This is why we implemented and compared with the "light-weight" version of transformers in our experiments (e.g., method (b) in Figure 2). The proposed CSPAN model, on the other hand, is more compact and approaches a satisfactory result in just a few epochs, and the time taken for each epoch is also much less than standard transformers. Therefore, our approach is computationally very efficient, especially for classification of short or median-length documents.

## 4 Conclusion

We presented the cascaded semantic and positional self-attention to aggregate *semantic* and *positional* information in document classification. It overcomes the limitation of existing positional encoding schemes, and shows encouraging performance against state-of-the-art methods using transformers and CNNs. In the meantime, it has a compact model size and is computational efficient. Our studies demonstrate the importance of properly aggregating semantic and positional components, and we will further extend it more challenging NLP tasks in our future research.

## Acknowledgments

Jie Zhang is supported by NSFC 61973086, Shanghai Municipal Science and Technology Major Project (No.2018SHZDZX01) and ZJ Lab.